\documentclass[journal]{IEEEtran}
\ifCLASSINFOpdf
\else
\fi
\usepackage{array, booktabs}
\usepackage{graphicx}
\usepackage{comment}
\usepackage{algorithm}
\usepackage{algorithmic}
\usepackage{bbding}
\usepackage{pifont}
\usepackage{amssymb}
\usepackage{booktabs}
\usepackage{comment}
\usepackage{bm}

\usepackage[hang,small,bf]{caption}
\usepackage[subrefformat=parens]{subcaption}
\usepackage{microtype}

\usepackage[normalem]{ulem}

\usepackage[cmex10]{amsmath}
\usepackage[capitalize]{cleveref}
\crefname{section}{Sec.}{Secs.}
\Crefname{section}{Section}{Sections}
\Crefname{table}{Table}{Tables}
\crefname{table}{Tab.}{Tabs.}

\usepackage{array}
\renewcommand{\baselinestretch}{0.99}

\usepackage{url}

\begin{document}
%
\title{PolMERLIN: Self-Supervised Polarimetric Complex SAR Image Despeckling with Masked Networks}
%
%
%

\author{Shunya~Kato, Masaki~Saito, Katsuhiko~Ishiguro, and Sol~Cummings\thanks{Shunya Kato is with the Graduate
School of Informatics, Kyoto University, Kyoto 606-8501, Japan (e-mail:
s-kato@nlp.ist.i.kyoto-u.ac.jp).}\thanks{Masaki Saito, Katsuhiko Ishiguro, and Sol Cummings are with Preferred Networks Inc., Tokyo 100-0004, Japan
(e-mail: msaito@preferred.jp; ishiguro@preferred.jp; scummings@preferred.jp).}}
\maketitle

\begin{abstract}
Despeckling is a crucial noise reduction task in improving the quality of synthetic aperture radar (SAR) images. Directly obtaining noise-free SAR images is a challenging task that has hindered the development of accurate despeckling algorithms. The advent of deep learning has facilitated the study of denoising models that learn from only noisy SAR images. However, existing methods deal solely with single-polarization images and cannot handle the multi-polarization images captured by modern satellites.
In this work, we present an extension of the existing model for generating single-polarization SAR images to handle multi-polarization SAR images. Specifically, we propose a novel self-supervised despeckling approach called channel masking, which exploits the relationship between polarizations. Additionally, we utilize a spatial masking method that addresses pixel-to-pixel correlations to further enhance the performance of our approach. By effectively incorporating multiple polarization information, our method surpasses current state-of-the-art methods in quantitative evaluation in both synthetic and real-world scenarios.
\end{abstract}


%
\IEEEpeerreviewmaketitle

\section{Introduction}
\label{sec:intro}

\begin{figure}[t]
\centering
\includegraphics[width=0.45\textwidth]{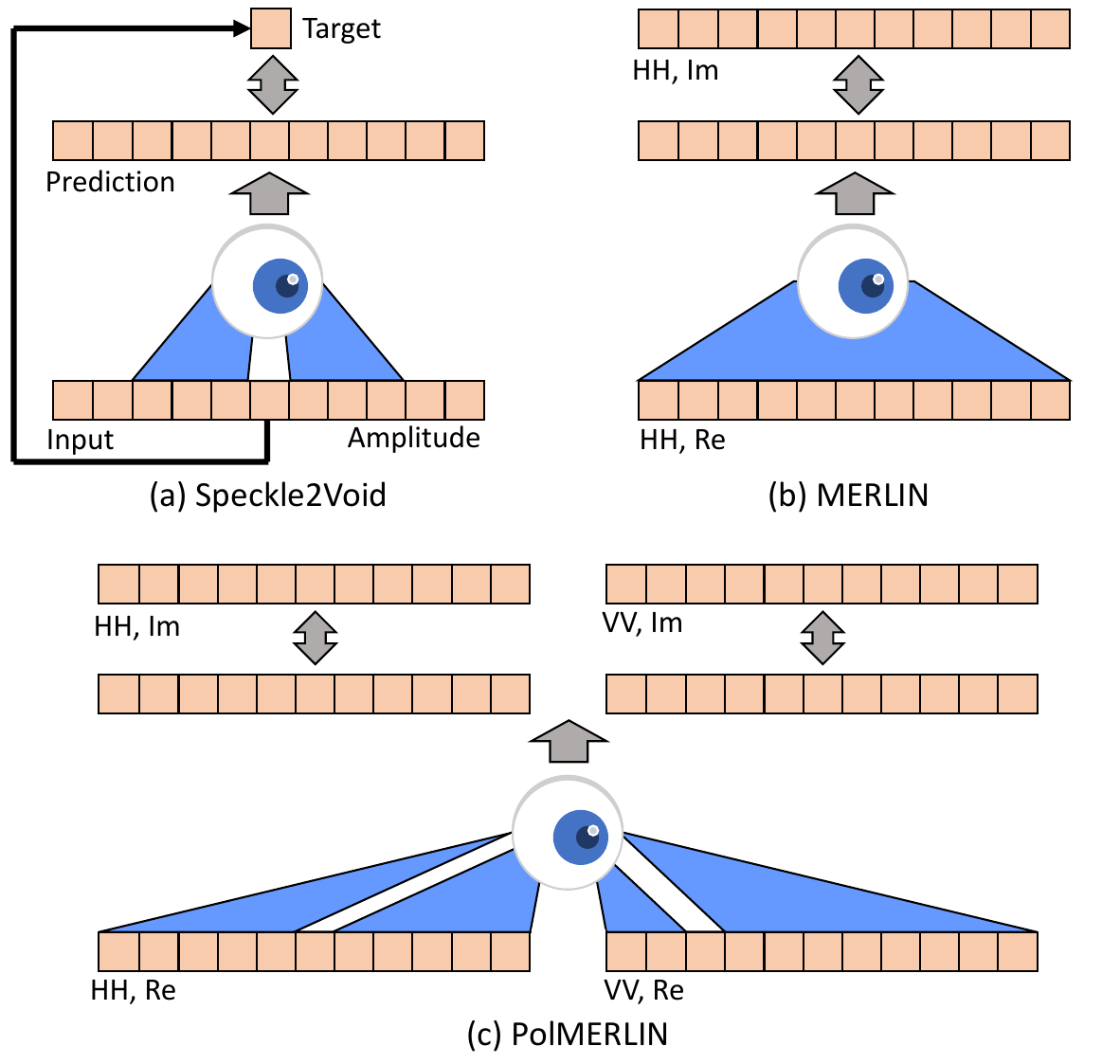}
\caption{
Comparison of self-supervised despeckling methods. Blue pixels indicate the range of receptive fields. (a) Speckle2Void uses blind-spotting (spatial information). (b) MERLIN predicts a Re/Im component from the counterpart component in a single polarization signal. (c) The proposed PolMERLIN predicts Re/Im components from the counterpart components of multi-polarization signals. To use spatial information as well, PolMERLIN intentionally removes some pixels from the receptive field.
}
\label{fig:fig1}
\vspace{-1em}
\end{figure}

SAR images suffer from signal-dependent and spatially correlated noise, which is commonly referred to as speckle noise.
This noise arises due to the interference of scattered waves, which are randomly phased, from numerous scattering surfaces. Due to the pronounced degradation of image quality by the speckle noise, despeckling of SAR images is crucial to preserve image fidelity and prevent performance degradation in downstream tasks, such as semantic segmentation~\cite{8647657}.

The tremendous strides made in deep learning have spurred the development of numerous despeckling techniques. These approaches generally fall into two categories of training methods: supervised and self-supervised. In the former~\cite{8128234,wang2017sar,perera2022transformer}, a network is trained from a set of paired noisy and noise-free images of the same subject, with the benefit of leveraging well-established denoising networks like CNN and Transformer. 
However, obtaining noise-free SAR images directly from satellites or aircraft remains a challenge \cite{Patra2021,Patra2023}. These studies circumvent this by calculating the mean of multi-temporal SAR images, which requires avoiding areas that have undergone changes, or by adding pseudo-speckle noise in non-remote sensing optical images, which creates a gap between training and testing conditions.~\cite{wang2017sar,perera2022transformer}.

On the other hand, self-supervised training can bypass this challenge, as it does not require noise-free SAR images.
Speckle2Void~\cite{molini2021speckle2void} addresses despeckling of SAR amplitude images using a self-supervised denoising method~\cite{krull2019noise2void}.
Additionally, MERLIN~\cite{dalsasso2021if} presents a self-supervised despeckling technique designed for two channels, based on the statistical property that the real and imaginary components of a complex SAR image are uncorrelated. However, this method is still limited to single polarization SAR images, thus constraining the available speckle noise data.

In this study, we present PolMERLIN, a novel self-supervised despeckling method that leverages ``multi-polarization complex SAR images'' to achieve superior performance. As depicted in Figure~\ref{fig:fig1}, PolMERLIN exploits the spatial correlation among multiple polarization complex images, in contrast to MERLIN, which can only handle single polarization complex SAR images. Specifically, we extend the statistical model of speckle noise from single-polarization complex SAR images to multi-polarization complex SAR images.

According to this model, we reveal that the same components between different polarizations are correlated, while different components are independent. 
Based on these findings concerning the (in)dependencies, we propose a novel self-supervised learning method, called {\it channel masking}, which masks the same component in a multi-polarization complex SAR image and predicts it from the other component.

While channel masking effectively despeckles by taking into account the independent relation between components, it overlooks the pixel-to-pixel relationship that is employed in~\cite{molini2021speckle2void}. To enhance the despeckling performance further, we incorporate the idea of spatial masking that masks a portion of the unmasked component~\cite{molini2021speckle2void,krull2019noise2void}.
By integrating these two techniques, PolMERLIN outperforms MERLIN, a state-of-the-art self-supervised despeckling method, in both simulated and real datasets.

\section{Statistical Model of Complex SAR}
\label{sec:statistical_model}
This section summarizes the generation process of single-polarization complex SAR images formulated in MERLIN~\cite{dalsasso2021if} as a basis of our generation model of multi-polarization complex SAR images.
It is based on the Speckle model of Goodman et al.~\cite{goodman2007speckle}, which states that radar reflection is much more sensitive to the roughness of the ground surface than the radar wavelength.
Let us denote a pixel value observed by the sensor as $z \in \mathbb{C}$. Then we can describe $z$ as the sum of independent reflected waves: $z_n \equiv \rho_n \exp (j \phi_n)$. 
That is, we can also represent $z = \sum_n \rho_n \exp (j \phi_n)$. 
Assuming that (i) the number of reflected waves is sufficiently large, (ii) amplitudes and phases are independent, and (iii) the phases distribute uniformly within ($-\pi$, $\pi$), then the distribution $p (z)$ of $z$ follows the following circularly symmetric complex Gaussian distribution:
\begin{equation}
\label{eq:p(z)}
    p (z) = \frac{1}{\pi r} \exp \left(- \frac{|z|^2}{r} \right),
\end{equation}
where $r > 0$ represents the reflectance of the SAR. That is, the variance of the noise observed by the sensor is determined by the magnitude of the SAR reflectance.
Decomposing $z$ into its real and imaginary components as $z = a + jb$, the distribution $p(z)$ can be expressed as $p(z) \propto \exp \left(- a^2 / r \right) \exp \left(- b^2 / r \right)$.
It shows that in $p(z)$ the distributions of the real and imaginary components are independent.

In the above equation, the joint distribution of the image, consisting of the array of values observed by the sensor, is assumed to be pixel-independent. However, the actual observed values are known to be spatially correlated due to small deformations originating from the sensor. To represent this, MERLIN uses a linear transformation of $z$ to represent the process of observing the actual SAR image.
Specifically, when we define the SAR image with $N$ pixels before the linear transformation as $\bm{z} \in \mathbb{C}^N$ and the linear transformation matrix as $\bm{T} \in \mathbb{R}^{N \times N}$, the spatially uncorrelated signal $\bm{z}$ can be transformed into a spatially correlated signal $\tilde{\bm{z}}$ represented as
$\tilde{\bm{z}} = \bm{Tz}$.
The real and imaginary components of z are similarly transformed and can be expressed as $\tilde{\bm{a}} \equiv \bm{Ta}$ and $\tilde{\bm{b}} \equiv \bm{Tb}$, respectively.
This transformation implies that although there is spatial correlation with respect to the real and imaginary component images of the actual complex SAR image, these distributions are independent as long as
$\bm{T}$ is a real matrix, i.e., $p(\tilde{\bm{z}})=p(\tilde{\bm{a}})p(\tilde{\bm{b}})$.

Using the above generation process, MERLIN performs self-supervised despeckling using the so-called masked training. Specifically, MERLIN performs masking on the complex input ($[\bm{a}, \bm{b}]$), for example, masking the real component ($\bm{a}$) and not masking the imaginary component ($\bm{b}$). It then trains a neural network to predict the real image from the imaginary image.
Dalsasso et. al. \cite{dalsasso2021if} have demonstrated that accurate despeckling can be achieved with these procedures.
\section{Proposed Method}
\label{sec:proposed_method}

\begin{figure*}[t]
\centering
\includegraphics[width=0.9\textwidth]{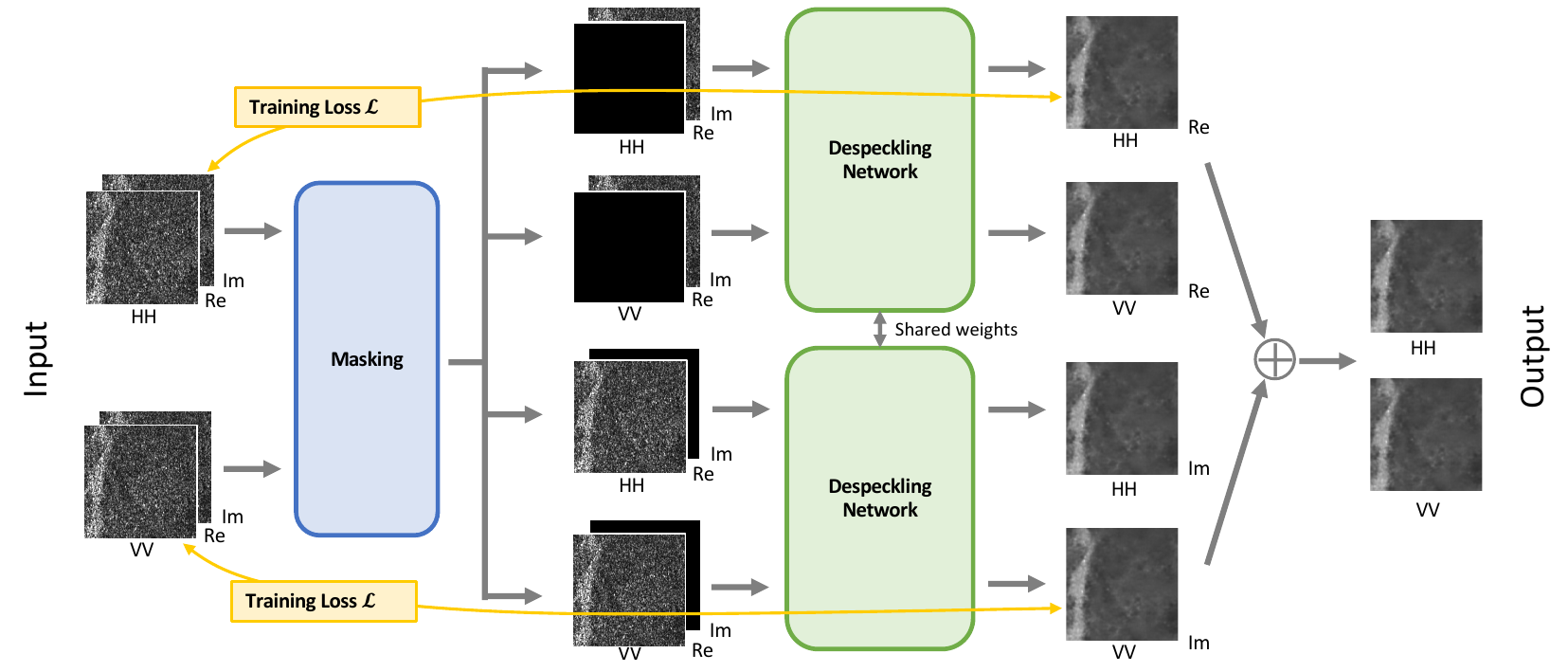}
\caption{
Overview of PolMERLIN. After channel masking and spatial masking are performed on the input, the model is trained by restoring the masked components using a despeckling network. For the despeckling network architecture we used the U-Net model~\cite{Wang2022}.}
\label{fig:overview}
\end{figure*}

\subsection{Statistical Model of Polarimetric Complex SAR}
\label{subsec:statistical_model_of_polarimetric_complex_sar}
To clarify the relationship between pixels in multi-polarization SAR images studied in this paper, we extend the statistical model of MERLIN (Sec.~\ref{sec:statistical_model}) to multi-polarization images. 
The complex amplitude $z$ is once defined as a scalar value in MERLIN. In our model, $z$ is extended as a vector, $\bm{\zeta}$, to capture multi-polarization. In this section, we assume $\bm{\zeta}$ as $\bm{\zeta} \equiv [z_{\mathrm{hh}} \ z_{\mathrm{vv}}]^T \in \mathbb{C}^2$ since the TerraSAR-X dataset used in our experiments is dual polarization data (HH, VV).
Using the stochastic model for the multi-polarization SAR~\cite{1237385}, 
the joint distribution $p(\bm{\zeta})$ is an extension of Equation~\ref{eq:p(z)} as follows:
\begin{equation}
\label{eq:p(eta)}
    p(\bm{\zeta}) = \frac{1}{\pi^2|\sum|}\exp(-\bm{\zeta}^H \Sigma^{-1} \bm{\zeta}),
\end{equation}
where $\Sigma$ is the Hermitian and covariance matrix defined as
$\Sigma = \mathbb{E}[\bm{\zeta}\bm{\zeta}^H]$,
where ${}^H$ represents the Hermitian operator.
Since $p(a_{\mathrm{hh}}) = p(b_{\mathrm{hh}})$ and $p(a_{\mathrm{vv}}) = p(b_{\mathrm{vv}})$ from the discussion in Sec.~\ref{sec:statistical_model},
one of the off-diagonal components of the above matrix can be written as $\mathrm{Im}(\mathbb{E}[z_{\mathrm{hh}}\overline{z_{\mathrm{vv}}}]) = \mathbb{E}[-a_{\mathrm{hh}}b_{\mathrm{vv}}+b_{\mathrm{hh}}a_{\mathrm{vv}}] = 0$.
Similarly, we can write $\mathrm{Im}(\mathbb{E}[\overline{z_{\mathrm{hh}}}z_{\mathrm{vv}}])=0$.
Therefore, $\Sigma$ can be redefined by the following real matrix: $\begin{bmatrix} r_{\mathrm{hh}} & r_{\mathrm{hv}} \\ r_{\mathrm{hv}} & r_{\mathrm{vv}} \end{bmatrix}$.
Note that from Equation~\ref{eq:p(z)}, $r_{\mathrm{hh}}$ and $r_{\mathrm{vv}}$ are the components of the SAR image we wish to predict.

From the above, the exponential component of Equation~\ref{eq:p(eta)} can be decomposed as
\begin{align}
    \bm{\zeta}^H \Sigma^{-1} \bm{\zeta} &= (r_{\mathrm{vv}}a_{\mathrm{hh}}{}^2- 2r_{\mathrm{hv}}a_{\mathrm{hh}}a_{\mathrm{vv}} + r_{\mathrm{hh}}a_{\mathrm{vv}}{}^2) \nonumber \\
    &+ (r_{\mathrm{vv}}b_{\mathrm{hh}}{}^2- 2r_{\mathrm{hv}}b_{\mathrm{hh}}b_{\mathrm{vv}} + r_{\mathrm{hh}}b_{\mathrm{vv}}{}^2).
\end{align}
Hence, if we represent $\bm{\alpha}$ and $\bm{\beta}$ by $\bm{\alpha} = [a_\mathrm{hh} \ a_\mathrm{vv}]^T$ and $\bm{\beta} = [b_\mathrm{hh} \ b_\mathrm{vv}]^T$, then $p(\bm{\zeta}) = p(\bm{\alpha})p(\bm{\beta})$ holds. Similarly, $p(\bm{\alpha}) \neq p(a_{\mathrm{hh}})p(a_{\mathrm{vv}})$ and $p(\bm{\beta}) \neq p(b_{\mathrm{hh}})p(b_{\mathrm{vv}})$ hold except in the obvious case of $r_{\mathrm{hv}} = 0$.
That is, $\bm{\alpha}$ and $\bm{\beta}$ are independent, but $a_{\mathrm{hh}}$ and $a_{\mathrm{vv}}$ are non-independent (the same applies to $b_{\mathrm{hh}}$ and $b_{\mathrm{vv}}$).

This means that not only is there independence between the real and imaginary parts of the same polarized image (e.g., $a_{\mathrm{vv}}$ and $a_{\mathrm{hh}}$) as assumed in MERLIN, but there is also similar independence between the real and imaginary parts of different polarized images (e.g., $a_{\mathrm{hh}}$ and $b_{\mathrm{vv}}$). It also means that assuming $\bm{T}$ is real, the real and imaginary components of any polarization images are still independent after the linear transformation.
On the other hand, since the real and imaginary components of different polarizations are not independent of each other, a simple extension of MERLIN (Noise2Noise~\cite{pmlr-v80-lehtinen18a} between arbitrary components) does not satisfy the above condition.

\subsection{Channel Masking}
\label{subsec:channel_masking}
Noise2Noise~\cite{pmlr-v80-lehtinen18a} enables denoising by training to recover another independent noisy image from an input, without explicitly defining ground truth or noise models.
MERLIN's approach~\cite{dalsasso2021if} is based on Noise2Noise.

Specifically, as shown in Sec.~\ref{sec:statistical_model}, according to the Goodman's speckle model, as long as $\bm{T}$ is a real square matrix, a single-polarization complex SAR image is considered to consist of two independent noisy images, one with a real component and the other with an imaginary component.
MERLIN uses them to generate noise-reduced SAR images from noisy SAR images by training the network to recover one element from the other.

The despeckling method for multi-polarization complex SAR images also follows Noise2Noise.
As shown in Sec.~\ref{subsec:statistical_model_of_polarimetric_complex_sar}, the real components of different polarizations are correlated with each other, and the real and imaginary components of different polarizations can be considered independent.
Based on this statistical model, the real and imaginary components of multiple polarizations can be regarded as a pair of images collapsed according to the same noise distribution.
Thus, we propose a method called channel masking, which masks one component of the multi-polarization complex SAR image and predicts the other component.
Figure~\ref{fig:overview} shows the procedure of the channel masking.
By defining multi-polarization complex SAR images and an input tensor after channel masking as $\bm{x} = [\bm{a}_{\mathrm{hh}} \ \bm{b}_{\mathrm{hh}} \ \bm{a}_{\mathrm{vv}} \ \bm{b}_{\mathrm{vv}}] \in \mathbb{R}^{N \times 4}$ and $\bm{x}_c$, respectively, we have $\bm{x}_c = \mathrm{M_c} \circ \bm{x}$,
where $\circ$ is a Hadamard product and $\mathrm{M_c} \in \{0, 1\}^{N \times 4}$ is a binary matrix with zero for the channel to be masked and one for all other channels.

\subsection{PolMERLIN}
\label{subsec:polmerlin}
Channel masking allows despeckling using only noisy SAR images when the data provided are multi-complex polarization images.
For further improvement, we introduce spatial masking: a training scheme that extends masking in the channel direction, as proposed in Noise2Void~\cite{krull2019noise2void}, to the spatial direction. 
Since speckle noise is spatially correlated, simply using only spatial masking violates the assumption of Noise2Void where noise is spatially independent.
Inspired by Noise2Void, Speckle2Void~\cite{molini2021speckle2void} uses several tricks to mitigate spatial correlation. However, this method does not solve the fundamental problem of spatial correlation because it is a pseudo-relaxation of spatial correlation and makes the network more complex. 
Thus, we have introduced a trick to mitigate the negative effects of spatial correlation by using two orthogonal pieces of information together: spatial masking and channel masking.
Specifically, the image $\bm{x}_s$ after spatial masking can be expressed as $\bm{x}_s = \mathrm{M_s} \circ \bm{x}$,
where $\mathrm{M_s} \in \{ 0, 1 \}^{N \times 4}$ is a binary matrix whose masked elements are zero and the rest are one, randomly determined with a certain probability.

Figure~\ref{fig:overview} shows an overview.
PolMERLIN uses channel masking to mask the channels representing either real or imaginary components to be predicted in a multi-polarization complex SAR image, and spatial masking to mask pixels in the channels that are not to be predicted.
The despeckling network recovers the elements masked by channel masking from the given tensor, i.e., $\bm{r} = f_\theta (\mathrm{M_c} \circ \mathrm{M_s} \circ \bm{x})$.
$f_\theta$ is the despeckling network and $\bm{r} = [\bm{r}_{\mathrm{hh}} \ \bm{r}_{\mathrm{vv}}] \in \mathbb{R}^{N \times 2}$.

At training time, the reconstructed components are compared with the original SAR image to evaluate how similar they are. The loss function $\mathcal{L}'$ is defined as an extension to that of MERLIN~\cite{dalsasso2021if} to multiple polarizations:
\begin{equation}
    \label{eq:merlin}
    \mathcal{L}' = \mathcal{L}(\bm{r}_{\mathrm{Re}}, \bm{x}_{\mathrm{Im}}) + \mathcal{L}(\bm{r}_{\mathrm{Im}}, \bm{x}_{\mathrm{Re}}),
\end{equation}
\begin{equation}
    \mathcal{L}(\bm{r}, \bm{\gamma}) = \sum_{k=1}^N \sum_{p \in \{\mathrm{hh}, \mathrm{vv}\}} \left( {\frac{1}{2}\mathrm{log}(r_{kp}) + \frac{\gamma_{kp}^2}{r_{kp}}} \right),
\end{equation}
where $\bm{r}_{\mathrm{Re}}$ and $\bm{r}_{\mathrm{Im}}$ are real and imaginary matrices of $\bm{r}$ in which the real or imaginary component of $\bm{x}$ is masked and the other is predicted for by $f_{\theta}$. $\bm{x}_\mathrm{Re}$ and $\bm{x}_\mathrm{Im}$ are the matrices from which the real and imaginary components of $\bm{x}$ are extracted.

The model's inference is carried out by feeding both the real and imaginary parts of the given noisy image into the Despeckling Network. Note that the computational cost of the proposed method is nearly the same as existing methods since there is almost no difference between the architectures of MERLIN and PolMERLIN. Actually, when the images were $4 \times 256 \times 256$ px, the costs for MERLIN and PolMERLIN were 36.96 GFlops and 37.51 GFlops, respectively.
\section{Experiments}
\label{sec:experiments}
In our experiments, we evaluated our method on two datasets: synthetic speckle noise images and real SAR images.

\subsection{Procedure}
\label{sec:procedure}
During testing, channel masking was applied to the real and imaginary components, respectively, and the average of the restored results was used as the final despeckling result defined by $\bm{r}'$, i.e., $\bm{r}' \equiv (\bm{r}_{\mathrm{Re}} + \bm{r}_{\mathrm{Im}}) / 2$.

\paragraph{Synthetic Speckle Noise}
In this experiment, 400 training and validation images of BSDS500~\cite{amfm_pami2011} were used for training, and 100 test images were used for evaluation. As for the synthetic speckle noise, we used noise that follows a gamma distribution with mean and variance of 1 as the equivalent of single-look speckle noise. 
We assume the image before applying the synthetic speckle noise as the ground truth. Then we can evaluate the performance of the model by comparing the GT with the image after denoising.
In this experiment, we adopt PSNR and SSIM as image quality evaluation metrics. Since optical images do not possess the concept of polarization, RGB channels were regarded as separate polarizations instead.
As there is also no complex component in optical images, the channels were replicated and considered as the real and imaginary components of each channel instead.

We chose two baseline methods: (i) a supervised method using the original ``clean'' images as the ground truth,
and (ii) MERLIN, a self-supervised method relying only on noisy images corrupted by the synthetic speckle noise.

\paragraph{Real SAR Images}
Following previous studies~\cite{molini2021speckle2void,dalsasso2021if}, we used TerraSAR-X imagery\footnote{\url{https://tpm-ds.eo.esa.int/oads/access/collection/TerraSAR-X/tree}} captured in StripMap mode with a spatial resolution of 3m as the dataset. Since PolMERLIN handles multi-polarization complex SAR images, only 8 SAR images with HH and VV polarization information and complex components were used in our experiments. Because of the huge size of the SAR images (average 16,000 $\times$ 16,000 px per image), we divided each image into $256 \times 256$ patches without overlap.
Finally, we used 30,654 patches for training and 2,640 patches for evaluation. Since there is no ground truth in this dataset, we used the Equivalent Number of Looks (ENL)~\cite{oliver2004understanding} as an evaluation metric, which is a statistical property of speckle noise commonly used in the despeckling domain. 
The higher the ENL value, the more successful the despeckle is. As a baseline for the comparison method, we used MERLIN~\cite{dalsasso2021if}, which is also a self-supervised training method.

\paragraph{Training Details}
For fair comparisons, all models used the same U-Net-like architecture~\cite{Wang2022}. 
We employed the AdamW optimizer with a learning rate of $10^{-5}$ and a batch size of 16.
All models were trained for 100 epochs in the experiments of the TerraSAR-X. In the BSD-500, our model was trained for 2400 epochs, while existing methods were trained for 800 epochs. This is due to the different number of epochs at which PSNR and SSIM converged.
To narrow the range of values, the inputs and outputs of all models were set to log scale, as in previous studies~\cite{dalsasso2021if}. 2\% of pixels were set to be masked in the spatial mask.
Equation~\ref{eq:merlin} was used for the loss of MERLIN and PolMERLIN, and an MSE loss was used for the supervised method. All models were trained on 8 Tesla V100 GPUs and PyTorch was used for implementation.

\subsection{Experiments on Synthetic Speckle Noise}
\begin{figure}[!t]
  \centering
  \tabcolsep 2pt
  \begin{tabular}{ccc}
      \begin{minipage}[b]{0.32\linewidth}
        \centering
        \includegraphics[width=\textwidth]{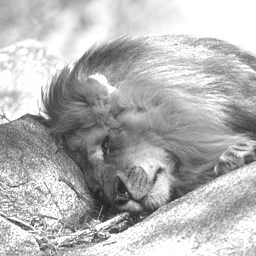}
        \subcaption{BSDS500}
      \end{minipage} &
      \begin{minipage}[b]{0.32\linewidth}
        \centering
        \includegraphics[width=\textwidth]{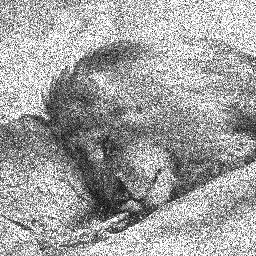}
        \subcaption{Noisy}
      \end{minipage} &
      \begin{minipage}[b]{0.32\linewidth}
        \centering
        \includegraphics[width=\textwidth]{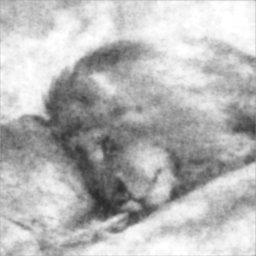}
        \subcaption{MERLIN}
      \end{minipage} \\
      \begin{minipage}[b]{0.32\linewidth}
        \centering
        \includegraphics[width=\textwidth]{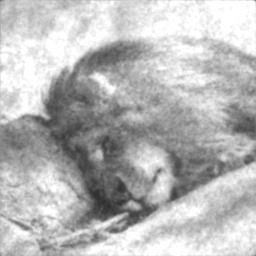}
        \subcaption{Ours (c)}
      \end{minipage} &
      \begin{minipage}[b]{0.32\linewidth}
        \centering
        \includegraphics[width=\textwidth]{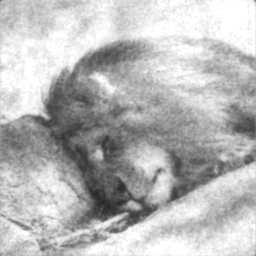}
        \subcaption{Ours (c+s)}
      \end{minipage}
  \end{tabular}
  \caption{
  Qualitative results for the R channel of the BSDS500 image despeckled with pseudo-noise. }
  \label{fig:quality of synthetic noise}
\end{figure}

\begin{figure*}[htbp]
  \centering
  \tabcolsep 2pt
  \begin{tabular}{ccc}
      \begin{minipage}[b]{0.32\linewidth}
        \centering
        \includegraphics[trim=0 6.0cm 0 4.0cm,clip,width=\textwidth]{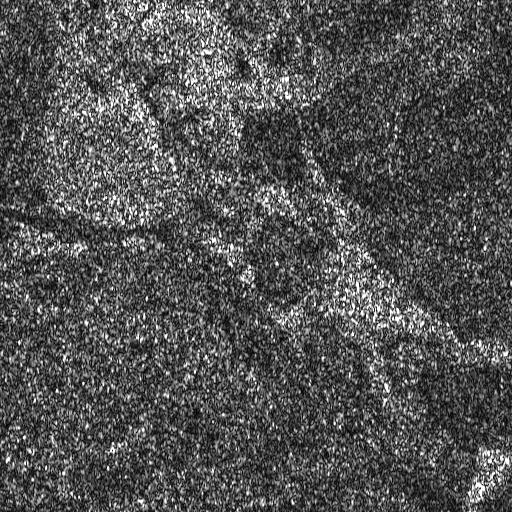}
        \subcaption{TerraSAR-X}
      \end{minipage} 
      \begin{minipage}[b]{0.32\linewidth}
        \centering
        \includegraphics[trim=0 6.0cm 0 4.0cm,clip,width=\textwidth]{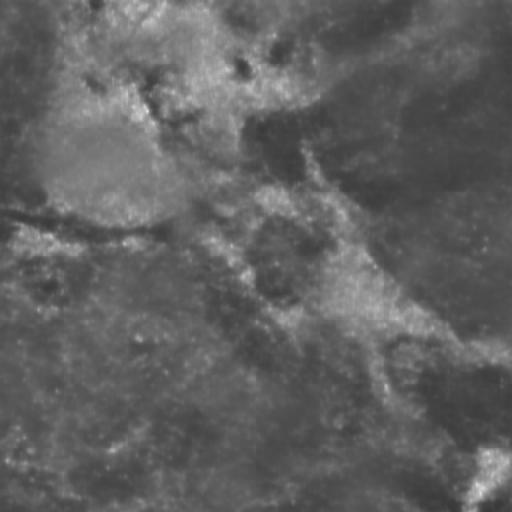}
        \subcaption{MERLIN}
      \end{minipage} 
      \begin{minipage}[b]{0.32\linewidth}
        \centering
        \includegraphics[trim=0 6.0cm 0 4.0cm,clip,width=\textwidth]{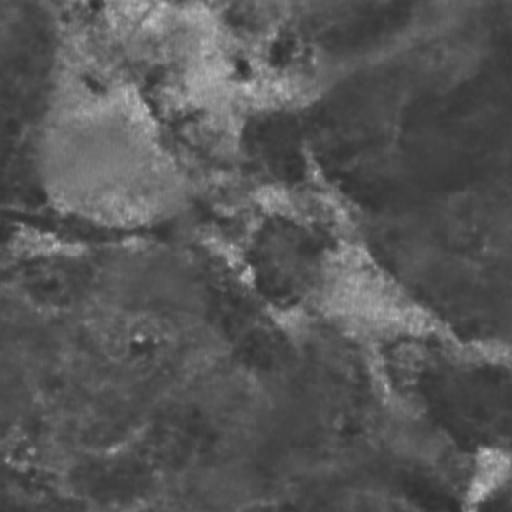}
        \subcaption{Ours (c+s)}
      \end{minipage} \\
  \end{tabular}
  \caption{
  Qualitative despeckling results of MERLIN and Ours(c+s) in TerraSAR-X.
Noisy bright (white) spots are found in the lower-left part of the MERLIN result. Ours(c+s)' result is not suffered from such artifacts. 
}
  \label{fig:quality of real speckle noise}
\end{figure*}

\begin{table}[t]
\centering
\small
\tabcolsep 2pt
\begin{tabular}{lccc}
\toprule
Method & R & G & B \\
\midrule
Input & $16.88/0.408$ & $16.69/0.367$ & $18.17/0.432$ \\
Supervised & $23.33/0.616$ & $23.69/0.634$ & $23.83/0.624$ \\
MERLIN & $23.30/0.617$ & $23.57/0.609$ & $23.51/0.610$ \\
Ours (c) & $23.61/0.668$ & $23.39/0.667$ & $23.30/0.620$ \\
Ours (c+s) & $\mathbf{24.14/0.687}$ & $\mathbf{24.50/0.696}$ & $\mathbf{24.01/0.646}$ \\
\bottomrule
\end{tabular}
\caption{Comparison of Despeckling results in terms of PSNR (left) and SSIM (right) on synthetic noise. ``Ours (c)'' is our model adapting only channel masking, whereas ``Ours (c+s)'' is our model adapting the both methods.}
\label{tab: quantitative analysis on synthetic noise}
\vspace{-1em}
\end{table}

Table~\ref{tab: quantitative analysis on synthetic noise} shows the quantitative results for synthetic noise. Compared to the input images, both despeckling methods show an improvement in PSNR and SSIM.
Comparing the MERLIN and the supervised method, we found that their PSNR and SSIM are almost the same. This indicates that the self-supervised, single-polarized MERLIN can perform as well as the supervised model.
Next, we compared Ours(c) with these two methods and found the Ours(c) model is evidently better than the other two in SSIM and is marginally better in PSNR. 
It implies that our method performs better despeckling than the existing methods.
We also compared Ours(c+s) with the other methods, and confirmed that it outperforms the others in both PSNR and SSIM, indicating that spatial masking contributes to despeckling improvement.
Figure~\ref{fig:quality of synthetic noise} shows the qualitative results of despeckling in synthetic noise. The results suggest that the image despeckled by MERLIN is blurred because only one channel is used, whereas our method uses the RGB channel and thus preserves the image detail relatively well.

\subsection{Experiments on Real SAR}

\begin{table}[t]
\centering
\small
\tabcolsep 2pt
\begin{tabular}{lcc||lcc}
\toprule
Method & HH & VV & Method & HH & VV \\
\midrule
Input & $2.9$ & $2.9$ & Ours (c) & $173.6$ & $191.8$ \\
MERLIN & $148.8$ & $186.3$ & Ours (c+s) & $\bm{274.0}$ & $\bm{301.3}$ \\ \hline
&&& mask HH/VV chs. & 38.0 & 134.4 \\
\bottomrule
\end{tabular}
\caption{Comparison of Despeckling Results in terms of ENL on Real SAR Images.
}
\label{tab: quantitative analysis on real sar image}
\vspace{-1em}
\end{table}

Table~\ref{tab: quantitative analysis on real sar image} shows the quantitative results for real speckle noise 
in terms of ENL.
It shows that all self-supervised despeckle methods significantly outperform the input images in both HH and VV, indicating that they are also capable of improving image quality by despeckling real SAR images.
A comparison with Ours(c) and MERLIN shows that our HH and VV results exceed those of MERLIN. 
This indicates that the use of multiple polarization images significantly improves despeckling performance in real SAR images.
In addition, Ours(c+s) outperforms Ours(c) by large margins in both HH and VV. 
It implies that adding spatial masking improves despeckling performance in real SAR images.
To check the validity of the probabilistic model discussed in Section~\ref{sec:proposed_method}, we also measured the ENL using a model (``mask HH/VV chs'') that predicts other images after masking either HH or VV images. The results are shown in Table~\ref{tab: quantitative analysis on real sar image}. These values are lower than MERLIN. This suggests a performance degradation due to information leakage, and at the same time, it indicates that our probabilistic model has a certain degree of validity.

Since the ENL only covers an aspect of the speckle noises, we cannot judge the quality of the despeckling based solely on the ENL values.
Therefore, we analyzed the qualitative results as well to determine the performance.
Figure~\ref{fig:quality of real speckle noise} shows the qualitative results of despeckling on real SAR images. 
We found that the proposed PolMERLIN successfully reduces noise while preserving semantic context without excessive smoothing, compared to MERLIN. 
These results illustrate the qualitative and quantitative performance improvements of our method over MERLIN.

\section{Conclusion}
\label{sec:conclusion}
We studied an efficient despeckling of multi-polarization complex SAR images. We first extended a known generative model of single-polarization complex SAR images to complex multi-polarization ones. Based on this model, we proposed a channel masking and a spatial masking method for self-supervised despeckling of multi-polarization complex SAR images.
Experiments showed that our method outperforms the despeckling method for single-polarization complex SAR images both quantitatively and qualitatively.
We are currently planning to test whether our proposed method is effective with other SAR wavelengths and airborne SAR as well. Additionally, since the training takes a long time to converge, establishing a method to shorten this duration is our future challenge.



\section*{Acknowledgment}
We would like to thank Shin-ichi Maeda and Tsukasa Takagi for helpful discussions.
This research work was financially supported by the Ministry of Internal Affairs and Communications of Japan
with a scheme of ``Research and development of advanced technologies for a user-adaptive remote sensing data platform'' (JPMI00316).

\ifCLASSOPTIONcaptionsoff
  \newpage
\fi



%
{\small
\bibliographystyle{ieee_fullname}
\bibliography{egbib}

\begin{thebibliography}{10}\itemsep=-1pt

\bibitem{amfm_pami2011}
Pablo Arbelaez, Michael Maire, Charless Fowlkes, and Jitendra Malik.
\newblock Contour detection and hierarchical image segmentation.
\newblock {\em IEEE Trans. Pattern Anal. Mach. Intell.}, 33(5):898--916, May
  2011.

\bibitem{8128234}
G. Chierchia, D. Cozzolino, G. Poggi, and L. Verdoliva.
\newblock Sar image despeckling through convolutional neural networks.
\newblock In {\em IGARSS}, 2017.

\bibitem{dalsasso2021if}
Emanuele Dalsasso, Lo{\"\i}c Denis, and Florence Tupin.
\newblock As if by magic: self-supervised training of deep despeckling networks
  with merlin.
\newblock {\em IEEE Transactions on Geoscience and Remote Sensing}, 60:1--13,
  2021.

\bibitem{8647657}
Yiping Duan, Xiaoming Tao, Chaoyi Han, Xiaowci Qin, and Jianhua Lu.
\newblock Multi-scale convolutional neural network for sar image semantic
  segmentation.
\newblock In {\em GLOBECOM}, 2018.

\bibitem{goodman2007speckle}
J.W. Goodman.
\newblock {\em Speckle Phenomena in Optics: Theory and Applications}.
\newblock Roberts \& Company, 2007.

\bibitem{krull2019noise2void}
Alexander Krull, Tim-Oliver Buchholz, and Florian Jug.
\newblock Noise2void-learning denoising from single noisy images.
\newblock In {\em Proceedings of the IEEE Conference on Computer Vision and
  Pattern Recognition}, pages 2129--2137, 2019.

\bibitem{pmlr-v80-lehtinen18a}
Jaakko Lehtinen, Jacob Munkberg, Jon Hasselgren, Samuli Laine, Tero Karras,
  Miika Aittala, and Timo Aila.
\newblock {N}oise2{N}oise: Learning image restoration without clean data.
\newblock In {\em ICML}, 2018.

\bibitem{1237385}
C. Lopez-Martinez and X. Fabregas.
\newblock Polarimetric sar speckle noise model.
\newblock {\em IEEE Transactions on Geoscience and Remote Sensing},
  41(10):2232--2242, 2003.

\bibitem{molini2021speckle2void}
Andrea~Bordone Molini, Diego Valsesia, Giulia Fracastoro, and Enrico Magli.
\newblock Speckle2void: Deep self-supervised sar despeckling with blind-spot
  convolutional neural networks.
\newblock {\em IEEE Transactions on Geoscience and Remote Sensing}, 60:1--17,
  2021.

\bibitem{oliver2004understanding}
C. Oliver and S. Quegan.
\newblock {\em Understanding Synthetic Aperture Radar Images}.
\newblock EngineeringPro collection. SciTech Publ., 2004.

\bibitem{Patra2021}
Anirban Patra, Arijit Saha, and Kallol Bhattacharya.
\newblock High-resolution image multiplexing using amplitude grating for remote
  sensing applications.
\newblock {\em Optical Engineering}, 60(7), 2018.

\bibitem{Patra2023}
Anirban Patra, Arijit Saha, and Kallol Bhattacharya.
\newblock Compression of high-resolution space video using phase grating.
\newblock {\em J Indian Soc Remote Sens}, 2023.

\bibitem{perera2022transformer}
Malsha~V Perera, Wele Gedara~Chaminda Bandara, Jeya Maria~Jose Valanarasu, and
  Vishal~M Patel.
\newblock Transformer-based sar image despeckling.
\newblock {\em arXiv preprint arXiv:2201.09355}, 2022.

\bibitem{wang2017sar}
Puyang Wang, He Zhang, and Vishal~M Patel.
\newblock Sar image despeckling using a convolutional neural network.
\newblock {\em IEEE Signal Processing Letters}, 24(12):1763--1767, 2017.

\bibitem{Wang2022}
Zejin Wang, Jiazheng Liu, Guoqing Li, and Hua Han.
\newblock Blind2unblind: Self-supervised image denoising with visible blind
  spots.
\newblock In {\em CVPR}, 2022.

\end{thebibliography}
}
%









\end{document}